\documentclass[letterpaper]{article} % DO NOT CHANGE THIS
\usepackage{aaa}  % DO NOT CHANGE THIS
\usepackage{times}  % DO NOT CHANGE THIS
\usepackage{helvet} % DO NOT CHANGE THIS
\usepackage{courier}  % DO NOT CHANGE THIS
\usepackage[hyphens]{url}  % DO NOT CHANGE THIS
\usepackage{graphicx} % DO NOT CHANGE THIS
\usepackage{natbib}  % DO NOT CHANGE THIS AND DO NOT ADD ANY OPTIONS TO IT
\usepackage{caption} % DO NOT CHANGE THIS AND DO NOT ADD ANY OPTIONS TO IT
\usepackage{booktabs}
\usepackage{multirow}
\title{ENAS4D: Efficient Multi-stage CNN Architecture Search for Dynamic Inference}
\author{
    Zhihang Yuan$ ^1$, 
    Xin Liu$ ^{2,3}$,
    Bingzhe Wu$ ^1$,
    Guangyu Sun$ ^1$
}
\affiliations{
    \textsuperscript{\rm 1}Peking University, Beijing, China\\
    \{yuanzhihang, wubingzhe, gsun\}@pku.edu.cn \\
    \textsuperscript{\rm 2} School of Computer Science and Technology, University of Chinese Academy of Sciences \\
    \textsuperscript{\rm 3}State Key Laboratory of Computer Architecture, Institute of Computing Technology, Chinese Academy of Sciences  \\
    liuxin19g@ict.ac.cn 
}

\begin{document}

\maketitle

\begin{abstract}
Dynamic inference is a feasible way to reduce the computational cost of convolutional neural network(CNN), which can dynamically adjust the computation for each input sample. 
One of the ways to achieve dynamic inference is to use multi-stage neural network, which contains a sub-network with prediction layer at each stage. The inference of a input sample can exit from early stage if the prediction of the stage is confident enough. 
However, design a multi-stage CNN architecture is a non-trivial task.
In this paper, we introduce a general framework, ENAS4D, which can efficiently search for optimal multi-stage CNN architecture for dynamic inference in a well-designed search space. 
Firstly,  we propose a method to construct the search space with multi-stage convolution. The search space include different numbers of layers, different kernel sizes and different numbers of channels for each stage and the resolution of input samples. 
Then, we train a once-for-all network that supports to sample diverse multi-stage CNN architecture. A specialized multi-stage network can be obtained from the once-for-all network without additional training.
Finally, we devise a method to efficiently search for the optimal multi-stage network that trades the accuracy off the computational cost taking the advantage of once-for-all network. 
The experiments on the ImageNet classification task demonstrate that the multi-stage CNNs searched by ENAS4D consistently outperform the state-of-the-art method for dyanmic inference. In particular, the network achieves 74.4\% ImageNet top-1 accuracy under 185M average MACs.
% 其中一种实现动态推理的方式是使用多阶段神经网络，这种网络包含多个前向预测阶段，每个阶段运行一个子网络。对于那些简单的样本，使用较少阶段以减少计算量，对于复杂样本，使用较多阶段保证准确率。
% 这篇文章中，我们提出了一种通过神经网络搜索来寻找多阶段神经网络的框架，ENAS4D。
% 利用once-for-all的特性
% 我们在分类任务上做了实验，ENAS4D，在ImageNet

% 我们提出一个灵活构建搜索空间的方法

\end{abstract}

\section{Introduction}

Deep convolutional neural networks~(CNNs) have demonstrated its powerful capabilities in computer vision tasks, such as image classification~\cite{He2016ResNet}, object detection~\cite{Ren2015fasterrcnn}, and image segmentation~\cite{DBLP:conf/iccv/MaskRCNN17}. However, the performance of CNNs always comes with a huge computational cost. Thus, various methods have been proposed to reduce the computational cost of the CNN inference, including efficient network
architecture design~\cite{Howard2017MobileNet}, network pruning~\cite{DBLP:conf/nips/WeightPrune15} and weight quantization~\cite{DBLP:conf/icml/GuptaAGN15}. 

Recently, dynamic inference has emerged as a promising way to reduce the computational cost by dynamically changing the network according to each input sample~\cite{Huang2018MSDNet,DBLP:conf/iclr/FBS19}. 
For example, we can allocate a small network for easy samples while a big network for hard ones in image classification task.
The small network is enough to classify the simple samples and it has a small computational cost. The big network with more computational cost is more capable to handle difficult samples than the small network. As a result, we can reduce the computational cost of easy samples while maintaining the overall prediction accuracy at the same time.

% 小网络足够分类简单样本，同时它的计算开销小

The multi-stage network is one of the solutions to achieve dynamic inference.
It has several inference stages, and each stage contains a sub-network with a prediction layer. 
At first, the input sample is sent to the first stage to perform the inference. 
If the confidence of the prediction is larger than a threshold, the inference process exits at this stage. In this paper, we use the highest value of the softmax output as the confidence.
If the confidence is smaller than the threshold, the inference of the next stage should be performed until the final stage is reached. 

% 一种常见的实现动态推理的方法是使用多阶段网络，cite。多阶段网络将神经网络分为几个推理阶段，每个推理阶段包含一个子神经网络和预测输出层。输入样本首先被送到第一个阶段的子网络进行前向推理，预测输出层输出网络的预测结果。如果输出预测结果的置信度大于设定的阈值，那么前向预测在这个阶段退出，该阶段的预测结果作为最终的预测结果。如果输出的预测结果的置信度小于设定的阈值，那么将神经网络送入下一个阶段进行进一步的预测，下一个阶段的神经网络可以利用上一个阶段的中间特征图像以减少计算量，得到结果后再将该阶段输出的置信度与阈值进行比较，确定是否进入下一个阶段。重复这个过程直到网络到达最后一个阶段。

However, design a multi-stage CNN architecture is a non-trivial task to meet the goal of maximizing prediction accuracy under the constraint of the computational cost.
Different from the static network with a single stage, we need to separately design the architecture for each stage.
At the same time, we are able to reuse the feature maps of previous stages by the latter stages in multi-stage network to avoid double calculate these features.
Therefore, designing a network in such a huge design space is very complicated.

% 目标是在限制平均计算开销的情况下尽可能提升准确率
% 不同于单阶段的静态网络
% 因此，在这样巨大的设计空间中设计网络很复杂，

% 设计优秀多阶段CNN的网络结构是一件困难的事情。设计多阶段网络的目标是在保证预测精度的情况下，使得平均计算开销最小。根据这个设计目标，我们需要为每一个阶段选择不同的网络结构、设置每一层的网络参数。同时，为了减少不同阶段之间重复的计算，不同阶段应当共享一些中间信息，也就是说，我们需要选择一个阶段的某些中间特征图送到后面的阶段使用。这就造成了一个巨大的设计空间，之前已经有大量工作设计了多种多阶段神经网络, cite。

In recent years, neural architecture search (NAS) has made great achievements in the design of neural networks~\cite{DBLP:conf/iclr/NAS17,Tan_EfficientNet_ICML19}, which can automatically searches for the optimal network architecture under certain constraints.
S2DNAS~\cite{Yuan_S2DNAS_arxiv2019} first propose to use NAS to search for the multi-stage CNN architectures. 
It transform a static networks to multi-stage networks by dividing each layer into several stages.
However, S2DNAS is largely limited to the original network architecture, which leads to a small search space and a limited performance.
Meanwhile, S2DNAS has to train all of the sampled architectures during the search process, which is too inefficient to apply to large tasks such as ImageNet classification.

% 最近几年，神经网络搜索在网络结构设计方面取得了丰硕的成果，在多个任务上使用NAS搜索到的网络的性能远远超过比人工设计的网络性能。cite。S2DNAS首先将NAS应用到多阶段神经网络的设计中来，它将一个静态的神经网络的每一层通过切分得到多阶段神经网络。然而，S2DNAS仅仅对如何拆分网络进行搜索，没有每一个阶段内部的网络设计进行搜索，因此它每一阶段都拥有相同的层数和卷积结构。同时，S2DNAS没有使用神经网络参数共享策略，导致在搜索过程中需要花费大量的时间训练每一个被采样到的网络样本，搜索效率很低。

% 它能自动地在一定限制条件下搜索最优的网络结构
% S2DNAS被很大程度上限制在原来网络的架构之中
% 太低效了以至于无法应用在分类ImageNet这样的大型任务上

\begin{figure}[tbp] % h:here 当前位置 % b bottom % t top % p 浮动
    \centering
    \includegraphics [width=0.9\linewidth]{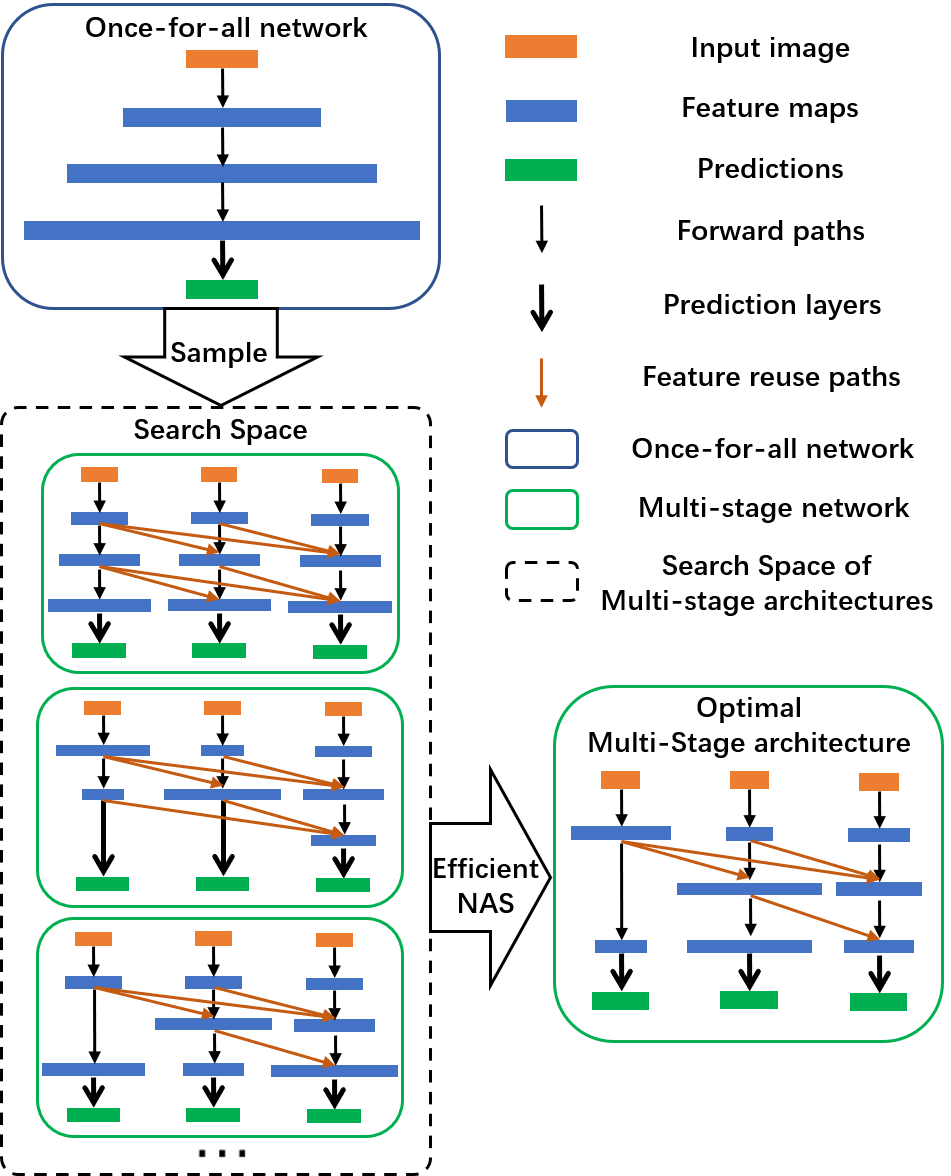}
    \caption{Overview of ENAS4D. We can sample sub-networks from the once-for-all network to build the search space of multi-stage architectures. Efficient NAS is used to search for the optimal multi-stage architecture.}
    \label{fig:overview}
\end{figure}

In this paper, we introduce a general framework, ENAS4D, which can efficiently search for optimal multi-stage CNN architecture for dynamic inference in a well-designed search space. 
Firstly,  we propose a method to construct the search space with multi-stage convolution. The search space include different numbers of layers, different kernel sizes and different numbers of channels for each stage and the resolution of input samples. The search space is significantly larger than the previous works
(about $10^{37}$ in our experiments).
Then, we train a once-for-all network that supports to sample diverse multi-stage CNN architecture. A specialized multi-stage network can be obtained from the once-for-all network without additional training.
Finally, we devise a method to efficiently search for the optimal multi-stage network that trades the accuracy off the computational cost taking the advantage of once-for-all network. Specifically, we devise a evaluation metric that combines the accuracy and computational cost of different stages to evaluate the multi-stage architecture. Then we train an metric predictor to predict the metric of a given architecture and we use  the evolutionary algorithm to search for the optimal multi-stage architecture. 
In this way, we can greatly reduces the cost of NAS.

% 搜索空间远大于之前的方法

% Next, we take the idea of once-for-all~\cite{Cai_onceforall_ICLR20} to train a large network that shares the parameters of different multi-stage CNN. 
% We can extract weights from the large network to set the weights of different multi-stage CNN. And this multi-stage CNN has good inference performance without retraining. 
% Therefore, we can quickly evaluate the performance of different multi-stage CNNs in the search space. 
% Finally, we train a performance predictor by sampling some multi-stage networks from the search space and evaluating their performance. 
% Using this predictor, we can efficiently search for the optimal multi-stage CNN architecture, which greatly reduces the cost of NAS.

% 这篇文章中，我们提出了一个通用的框架ENAS4D，它包含一个很大的搜索空间并能在这个空间内高效地搜索多阶段CNN。首先，我们提出了一个多阶段网络的搜索空间，每个多阶段网络的每个阶段都可以有不同的网络层数，每个卷积层可以有不同的卷积核大小和卷积通道数目，同时网络的输入分辨率也作为搜索的参数。接下来，我们利用once-for-all的想法，提出训练一个大网络来共享多阶段神经网络参数的方法。我们能从这个大网络中抽取一部分参数来设置不同的多阶段网络的网络参数，这个多阶段网络不需要重新训练就有很好的预测表现。因此我们能够快速地评估搜索空间中不同的多阶段神经网络的性能。最后，利用这个特点，我们从搜索空间中采样一些多阶段网络来训练一个性能预测器。利用这个预测器，我们高效地神经网络结构搜索最优的多阶段神经网络，大大减小了NAS的开销。

Experiments on image classification task shows the multi-stage CNN generated by ENAS4D consistently outperforms state-of-the-art methods for dynamic inference. In particular, the network achieves 74.4\% ImageNet top-1 accuracy under 185M average MACs. The time cost is about 2200 GPU hours in total and we can derive new specialized neural networks for many different constraint in minuets.

\section{Related Work}

\subsection{Dynamic Inference}
The idea of dynamic inference is to allocate different computation for different input sample and it is also called adaptive inference.
Some of the previous works attempted to add controllers to select which computations are executed.
For example, 
\cite{DBLP:conf/cvpr/LCCL17} skip the computation of inactive pixels in feature maps by using a extra convolution layers as the controller;
\cite{Lin2017RNP,DBLP:conf/iclr/FBS19,WangDynamicPruning2020} skip the computation of a set of channels in convolution based on the attention generated by additional linear layers or RNN;
\cite{Liu2018TradeOff,Wu2018BlockDrop,DBLP:conf/eccv/ConvnetAIG18,Wang2018SkipNet} proposed to dynamically drop the whole layers or blocks;
\cite{ChenDynamicConvolution2020cvpr} dynamically aggregates multiple parallel convolution kernels based on the kernel-wise attention.

On the other hand, multi-stage networks is a feasible way to achieve dynamic inference, which contain multiple stages. The inference of input sample can exit from any stage according to the confidence for the prediction of different stages. 
\cite{DBLP:conf/date/PandaSR16,Teerapittayanon2016BranchyNet,DBLP:conf/icann/CascadedInference19a} augment the static CNNs with additional side branch classifiers to enable dynamic inference.
\cite{Huang2018MSDNet} design a novel multi-stage network architecture to reuse the feature maps of different stages by inter-connecting them with dense connectivity.
\cite{YangRANetCVPR20} use the low-resolution representations for classifying “easy” inputs by designing the resolution adaptive networks.
Recently, \cite{Yuan_S2DNAS_arxiv2019} propose to transform a given static network to a multi-stage network to support dynamic inference by using network architecture search.

% initial controller选择、动态选择（DY-Net）、多阶段网络
% 

\subsection{Neural Architecture Search}
Neural architecture search~(NAS) is to automatically design the network architecture.
\cite{DBLP:conf/iclr/NAS17} use reinforcement learning to sample architectures. 
However, the sampled network should be train form scratch and this takes too much time and the computational cost is not considered. 
\cite{Pham_ENAS_ICML2018,Liu_DARTS_ICLR2019} share the weights among different architectures to boost the search efficiency.
\cite{Tan2018MnasNet,Tan_EfficientNet_ICML19,Cai_onceforall_ICLR20} combine the accuracy of network and computational cost of the hardware into the search target.

\section{Method}

\subsection{Overview of ENAS4D}
The overview of ENAS4D is depicted in Figure~\ref{fig:overview}. Firstly, a once-for-all network is trained. We can sample sub-networks from it with different settings, including different numbers of layers, kernel sizes, channels of intermediate feature maps and input resolution.
Next, We use these sub-networks to build the multi-stage CNN that can support dynamic inference. These multi-stage CNN architectures compose the search space. 
Finally, an efficient NAS algorithm is used to search for the optimal multi-stage architecture.

We first introduce the structure of the multi-stage CNN in the search space; then how to generate them with once-for-all network; last the search algorithm.

% , which is used to generate a search space comprises of dynamic models based on a given static CNN model.
% Specifically, we define two transformations and then apply the transformations to the original model for generating different dynamic models in the search space. Each of these dynamic models is a multi-stage CNN that can be directly used for dynamic inference. All these generated models form the search space. Once the search space is generated, {\tt NAS} searches for the optimal model in the space. In what follows, we will give the details of these two components.

% 我们使用这些子网络来构建多阶段神经网络，这些多阶段神经网络能进行动态推理。接下来，我们将这些多阶段神经网络组成搜索空间。最后，我们使用高效的NAS算法来搜索其中最优秀的一个多阶段神经网络。

\subsection{The structure of Multi-stage CNN}
Multi-stage CNN is one of the methods to achieve dynamic inference.
It has $S$ inference stages, and each stage contains a sub-network $f_s$ with a prediction layer.
An input sample $x$ is sent to the network to perform the inference stage by stage. 
If the confidence of the prediction of the stage $s$ is larger than a threshold $T_s$, the inference process exits at this stage.
Otherwise, the inference of the next stage should be performed until the final stage is reached. 

In this way, it can dynamically change the computation cost for different input samples.
When designing the multi-stage CNN architectures, we can use different network architectures for each sub-network to maximize the predictive performance of different stages. 
Moreover, we can reuse the feature maps generated by the previous stage in the later stages, thereby further reducing the computation cost of the sub-network in the later stages. 

In the following part, we will introduce the basic component called multi-stage convolution, then how to use it to build multi-stage residual block and multi-stage network.
% Specifically, we save the intermediate feature map calculated in the previous stage, and use it when calculating the new feature map in the later stage.

% multi-stage CNN 是实现动态推理的方法之一，其包含多个阶段的子网络。输入图片首先被送到第一个阶段的子网络中进行前向推理，第一个阶段的预测层输出预测结果。如果预测的置信度高于阈值，前向预测从这里退出；如果预测的置信度低于阈值，图片将送入第二个子网络进行预测。我们可以为每个子网络设计不同的网络结构，以最大化网络在不同阶段的预测性能。同时，我们还可以在不同阶段之间传递信息，以重复利用前面阶段计算得到的特征，从而减少后面阶段子神经网络的计算开销。具体来说，我们将前面阶段计算出的中间特征图保存下来，并在后面阶段计算新的特征图的时候被用到。

% 接下来，我们首先介绍多阶段网络的基本组成单元，然后介绍如果用这样的基本单元组成多阶段网络，最后介绍如果从once-for-all网络中采样得到这样的多阶段网络。

\subsubsection{Multi-stage Convolution}

\begin{figure}[tbp] % h:here 当前位置 % b bottom % t top % p 浮动
    \centering
    \includegraphics [width=0.75\linewidth]{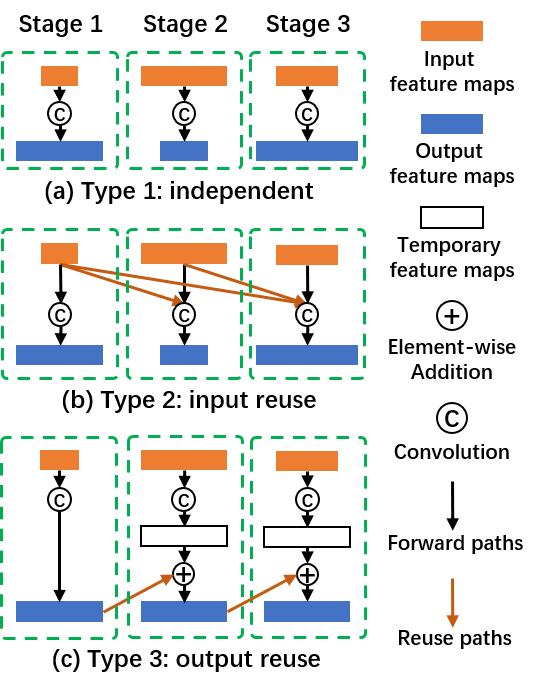}
    \caption{Three types of multi-stage convolution. The length of the square represents the number of channels of the feature maps.}
    \label{fig:multi-stage_convolution}
\end{figure}

% 方块的长度代表特征图的channel数量

The convolution layer is the most important component of CNN. It processes $C$ input feature maps $X=\{X_1,X_2,\dots,X_C\}$ and outputs $O$ output feature maps $Y=\{Y_1,Y_2,\dots,Y_O\}$ with weight $W\in R^{O\times C \times K \times K}$, where $K$ is the kernel size. The typical convolution is formulated as:
$$Y_j=\sum_{i=1}^C{\rm CONV}(W_{ji}, X_i),$$
where the ${\rm CONV}$ is the basic convolution operation on a input channel using a convolution filter.

In order to support dynamic inference, we design three types of multi-stage convolution, namely independent, input reuse and output reuse. The first type is called \textbf{independent} that is to use $S$ independent convolutions as shown in Figure~\ref{fig:multi-stage_convolution}(a). For the input of each stage, the weight of the corresponding stage is selected for convolution operation, which is formulated as:
$$Y^s_{j}=\sum_{i=1}^{C_s}{\rm CONV} (W_{ji}^{ss}, X^s_i), $$
where the $Y^{s}$ is the output of stage $s$, the $X^{s}$ is the input of stage $s$, the $W^{ss}$ is the weight of convolution in stage $s$ and $C_s$ is the number of input channels of stage $s$.
We can choose different numbers of output channels and different kernel sizes for different stages for this type of multi-stage convolution. 

The second is \textbf{input reuse}, which is shown in Figure~\ref{fig:multi-stage_convolution}(b). 
In order to reuse the input feature maps of the previous stage, we concatenate them with the input of the current stage as the input of the convolution, which is formulated as:
$$Y^s_{j}=\sum_{k=1}^{s}\sum_{i=1}^{C_k}{\rm CONV} (W_{ji}^{ks}, X^k_i), $$
where the $W^{ks}$ is the part of weight for $X^{k}$ and $Y^{s}$.

The third type is called \textbf{output reuse} and it is shown in Figure~\ref{fig:multi-stage_convolution}(c). The output of the convolution in each stage will be element-wisely added to the result of the previous stage. In order to use this type of multi-stage convolution, the number of output channels by all stages needs to be the same. This type of multi-stage convolution is formulated as:
$$Y^s_{j}=\sum_{i=1}^{C_s}{\rm CONV} (W_{ji}^{ss}, X^s_i)+Y^{s-1}_{j}.$$

% 卷积层是CNN最重要的主要组成成分，它对C个输入特征图进行处理，输出O个输出特征图。卷积操作可以表示为。其中w是卷积的权重（是卷积核）。

% 多阶段的卷积有两种实现方式。 两种方式对输出的处理不同，第一种方式是使用$S$个正常的卷积，每个阶段的卷积有单独的输出，它们可以选择不同的输出特征图通道数。为了重复利用特征图，我们将前面阶段的每所有输入作为后面阶段的输入。表示为：$Y^_s_{j}=\sum_k\sum_i(\rm{CONV}(W_ij^{jj}))$

% 每一个阶段的卷积输出会和上一个阶段的结果累积加到一起。为了使用这种类型的多阶段卷积，所有阶段输出的特征图通道数需要保持相同。
% 在一些情况下，我们可以。对每个阶段的输入，相应的阶段的权重被选用进行卷积操作。

\subsubsection{Multi-stage Residual Block} 

\begin{figure}[tbp] % h:here 当前位置 % b bottom % t top % p 浮动
    \centering
    \includegraphics [width=0.9\linewidth]{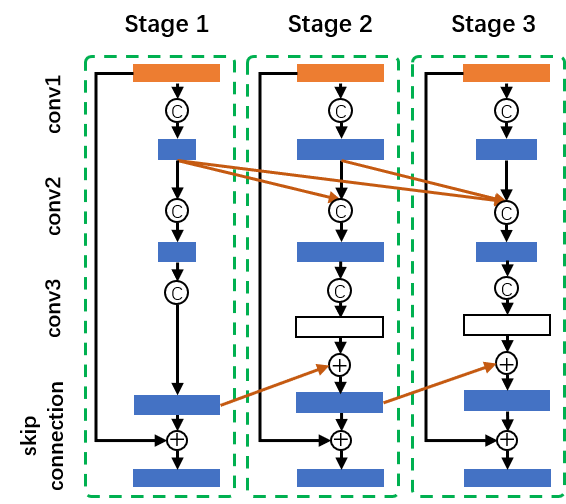}
    \caption{An example of multi-stage residual block, which contains three multi-stage convolutions in independent type, input reuse type and output reuse type respectively.}
    \label{fig:multi-stage_block}
\end{figure}

The residual block is proposed in~\cite{He2016ResNet}, which is widely used in a lot of CNN architectures. A residual block contains several convolution layers which process the input feature maps and a residual connection is used to add up the input and convolution output.
We designed a method to build the multi-stage residual block using the multi-stage convolution.
As demonstrated in Figure~\ref{fig:multi-stage_block}, the first convolution layer of the residual block uses independent type and the last convolution uses output reuse type, which ensures the channels of input feature maps and that of the output feature maps is consistent at each stage. 
We can assign different types for the other convolution layers in different residual blocks. For example, input reuse type is used for the second layer in the figure, and independent type is used for the depth-wise convolution in our experiment.

% 残差块被大规模应用在卷积神经网络中，残差块包含几层卷积处理输入，并将卷积的输出与残差块的输入相加作为残差块的输出。如图所示，我们针对各种不同的设计了通用的多阶段的残差块。残差连接被用于把输入和卷积输出加起来。

% 残差块的第一层的卷积使用独立型的多阶段卷积。最后一层的卷积使用输入输出复用型的多阶段卷积，这样保证了每个阶段残差块的输出特征图的通道数一致。其它卷积层根据可以根据需要使用不同的类型，如图我们在MobileNet的残差块的中间层使用独立型卷积。

\subsubsection{Multi-stage Network}

\begin{figure}[tbp] % h:here 当前位置 % b bottom % t top % p 浮动
    \centering
    \includegraphics [width=0.9\linewidth]{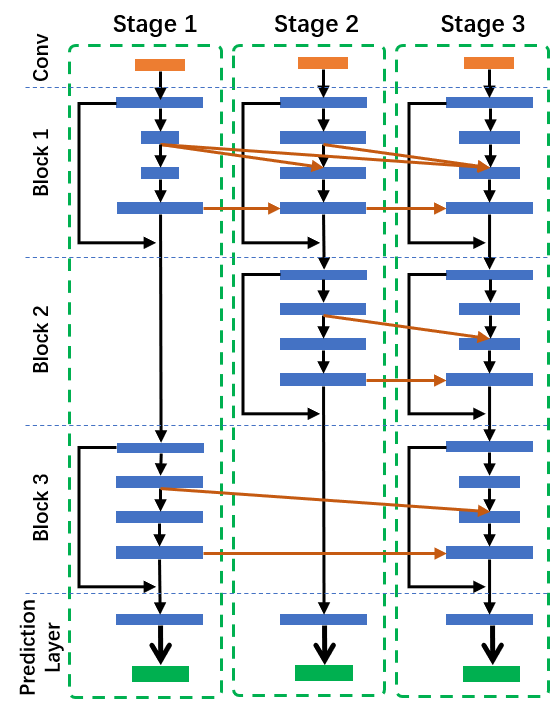}
    \caption{An example of multi-stage network containing three multi-stage residual blocks. We omit the convolution, temporary feature maps and element-wise addition.}
    \label{fig:multi-stage_network}
\end{figure}

Next, we introduce how to build a multi-stage network by using multi-stage convolution and multi-stage block. 
As shown in the Figure~\ref{fig:multi-stage_network}, the network contains a multi-stage convolution of independent type as the first layer and three multi-stage residual blocks. The prediction layer is added to the end of each stage. 
Note that we can skip the computation of some layers in some stages, as an example, the third block skips the second stage in the figure. In this way, we can set different numbers of layers for different stages of the network.

% 接下来，我们介绍如何构建多阶段网络。如图所示，我们将卷积层和残差块拼起来，形成多阶段网络。注意，每一个block可以选择在一些阶段跳过计算，如图中第三个block跳过了第二个阶段的计算。结果是，我们可以为不同阶段的网络设置不同的网络层数

\subsection{Sample Multi-stage CNN from Once-for-all Network}

An once-for-all network consists a few of convolution layers and residual block groups. A residual group contains $N$ residual blocks and each residual blocks contains several convolution layers. Some of the convolution layers can be stride of 2 to reduce the resolution of the feature map.
Our goal is to sample different multi-stage CNN architectures from the once-for-all network, including different number of layers or blocks in each stage, different kernel size in each layer, different number of channels in each stage, and different input image resolution. 
Next, we will introduce how to sample with different configurations.

% once-for-all网络的目标是能从中抽样出来不同的多阶段神经网络，包括四中不同的架构维度，分别是：每个阶段的网络层数、卷积输出通道数量、卷积核大小、网络输入图片分辨率。和一样，我们以残差块的组为基本单元构建once-for-all网络，每个组里面第一个卷积使用stride为2的卷积来减少特征图的分辨率。

% 我们可以从每个残差块的组中位不同阶段抽样出不同数量的残差块，每一个残差块中可以使用不同数量的中间层通道数和卷积核大小。另外，once-for-all网络能接受任意大小的输入图片大小。接下来将介绍具体如何抽样。

\subsubsection{Sample multi-stage width}

% \begin{figure}[tbp] % h:here 当前位置 % b bottom % t top % p 浮动
%     \centering
%     \includegraphics [width=0.9\linewidth]{images/multi-stage_width.png}
%     \caption{Demonstration of multi-stage network. Each block contains two multi-stage convolutions.}
%     \label{fig:multi-stage_network}
% \end{figure}

Width means the number of channels.
For single convolution layer, we first sort the channels based on the L1 norm of weight in once-for-all network. In order to generate $\{C_1,C_2,\dots,C_S\}$ number of channels for $S$ stages for independent type and input reuse type, we assign the most important $C_1$ channels to stage $1$, next $C_2$ channels to stage $2$ and so on. Thereby, the earlier stages use the more important channels to ensure accuracy. For the output reuse type, we use all of the output channels of the layer in each stage. The weight is initialized as the weight for corresponding input and output channels.
For residual blocks, since the output reuse type is used for the last convolution, we should sample the channels in each stage for the intermediate layers.
% The width is sampled from the width pool of different layers and the sum of $C$ is  
%For the last convolution, we didn't divide the output channels in once-for-all network. The output channels are shared by different stages.

% 由于每个block的所有阶段的输入输出维度被限制为相等，我们仅需要抽样中间层的通道数量
% 越前面的阶段使用越重要的通道来保证准确
% 输出网络层被不同的阶段共享
% 一定比例的

\subsubsection{Sample multi-stage kernel size}
The same as~\cite{Cai_onceforall_ICLR20}, we use the center of a 7x7 convolution kernel in once-for-all network to multiply with a transformation matrix to get the 5x5 kernel and the center of the 5x5 kernel to multiply with another transformation matrix to get the 3x3 kernel.
We sample different kernel sizes for different multi-stage convolution layers and initialize them with the weight after transformation in the corresponding location.
% In order to generate $\{K_1,K_2,\dots,K_S\}$ kernel sizes for $S$ stages of the multi-stage convolution, we transform the corresponding 7x7 kernel of once-for-all network to $K_s$x$K_s$ in stage $s$. 

\subsubsection{Sample multi-stage depth}
In order to generate $\{D_1,D_2,\dots,D_S\}$ numbers of blocks for $S$ stages in a block group, we keep the first $D_s$ blocks in stage $s$. The stage $s$ skip the computation of the last $N-D_s$ blocks. If the stage $s$ is skipped, the next stages $s+1$ need to calculate the channels of the stage $s$ to ensure important channels are not skipped. $D_s$ is sampled from a depth pool and we constraint $D_i\le D_j$ when $i<j$.

% 如果一层的前面阶段被跳过，那么后面阶段的channel需要融合前面阶段的channel以保证重要的channel计算不被跳过

\subsubsection{Sample the resolution of input image}
In order to keep the resolution of input feature maps in each stage equal to the resolution of the reused feature maps, we share the resolution of input image $R$ at each stage. The resolution is sampled from a image resolution pool.

% 我们共享每个阶段输入图片的分辨率
% 使得每个阶段输入的分辨率与复用的特征图分辨率相等
\subsection{Training the Once-for-all Network}

% We also use the method of progressively shrinking~\cite{} to train the once-for-all network. 
We sample multi-stage sub-networks to train at each training iteration. 
Specifically, after a multi-stage sub-network is sampled, inference is performed for each stage and the output of the prediction layer in each stage is got.
Then, we calculate the loss for each stage separately, and sum these losses to get the total loss. Finally, we perform back propagation with the total loss and update the weight. 
In order to improve the performance, knowledge distillation is used for the loss at each stage to mimic the output of complete once-for-all network.
% 我们也使用progressively shringking的方法，对once-for-all进行训练。和cite不同的是，我们在训练阶段，每次采样的样本是一个多阶段CNN。多阶段CNN有多个预测层的输出。我们分别计算每一个阶段输出的loss，然后将这些loss加起来得到总loss进行反向传播。同样，我们也是用知识蒸馏的方法到每一个阶段输出。

\subsection{Efficient Network Architecture Search}

After training of the once-for-all network, NAS is be used to search for the optimal multi-stage architecture with high accuracy and low computational cost. 
Firstly, we devise a evaluation metric $R$ combining the accuracy and calculation cost of different stages to evaluate the multi-stage architecture. Then, we train an metric predictor to predict the metric of a model given its architecture and input image size. Finally, we use the evolutionary algorithm to search for the optimal multi-stage architecture.

% 在训练完成once-for-all网络之后，接下来需要使用NAS来搜索最优的多阶段架构with high accuracy and low computational cost。

\subsubsection{Evaluation metric}
A dataset $\mathcal{D}$ is used to evaluate the metric on a multi-stage CNN.
Using dynamic inference, each input sample in a dataset $\mathcal{D}$ exit from different stages, and we count the number of samples exit from each stage (notated as $N_s$) and the prediction accuracy of each stage (notated as ${\rm ACC}_s$). We can also count the accumulative computational cost (MACs or latency on specific device) of the inference exit from each stage (notated as ${\rm COST}_s$). Then we calculate the average accuracy and the average computational cost over the dataset:
$${\rm ACC}_{avg}=\frac{1}{|\mathcal{D}|}\sum_{s=1}^S {\rm ACC}_s\times N_s$$ 
$${\rm COST}_{avg}=\frac{1}{|\mathcal{D}|}\sum_{s=1}^S {\rm COST}_s\times N_s,$$
where $|\mathcal{D}|$ is the number of samples in the dataset. Then we define the metric $R$ by combining them:
$$R= {\rm ACC_{avg}}\times ({\rm MIN(\frac{\rm COST_{target}}{\rm COST_{avg}} ,1)})^{\omega},$$
where ${\rm COST_{target}}$ is the target cost and the factor $\omega$ is used to adjust the relative importance between accuracy and computational cost.

% 为了评价从once-for-all采样出来的架构，我们使用了一个结合节准确率和计算代价综合指标：
% 使用动态推理我们能获得每一个样本在哪个阶段退出，并且能计算出每个阶段退出的样本的分类准确率。
% 我们将阶段s的预测准确率记作

\subsubsection{Grid search for threshold}

Firstly, we set the threshold of each stage to 1. 
Then, the multi-stage network inference the samples in $\mathcal{D}$  without early exiting. We obtain the prediction confidence and correctness of each sample in each stage, and record them in a database. 
At last, given $\omega$ and ${\rm COST_{target}}$, the grid search is used to search for thresholds for all stages to maximize the evaluation metric. 
The time cost of grid search is negligible (about 200ms in our experiment). 
Therefore, after the confidence and accuracy database is established, we can quickly set thresholds for different $\omega$   ${\rm COST_{target}}$.

% 首先，我们将数据集中的样本输入到所有阶段中进行预测，得到每一个样本在每一个阶段的预测置信度和正确性，并记录到数据库中。接下来，给定\omega和COST_target，使用网格搜索的方法搜索在evaluation metric最大的阈值。网格搜索的时间几乎可以忽略不计（我们的实验中为200ms左右）。因此在置信度和正确率数据库建立之后，我们可以快速地对针对不同的\omega和COST_target筛选最优阈值。

\subsubsection{Metric predictor}

Since it takes a long time to evaluate a network using the test dataset, we train a metric predictor which is a multi-layer perceptron to predict the metric of a model given its architecture and input image size. Firstly, we randomly sample 18K multi-stage sub-networks with different architectures and input image sizes, then evaluate the metric on 10K images which is sampled from the training dataset.
Then, we encode the architecture settings into the a feature vector by concatenate the one-hot vector of widths, kernel sizes, depths and input resolution.
At last, we train the predictor given the feature vector as input and the metric as target. 

% 由于使用测试集评价一个网络的需要花较长时间，我们训练一个metric的预测器 which is a three-layer perceptron to predict the metric of a model given its architecture and input image size。 

\subsubsection{Search with evolutionary algorithm}
We can quickly generate multi-stage architectures and test the predicted the metric using trained metric predictor. 
Evolutionary algorithm~\cite{Real_EvolutionSearch_AAAI2019} is used to search the optimal multi-stage architecture. 
In our experiment, the search process takes only several minutes.

\section{Experiments}
\label{sec:experiments}

To verify the effectiveness of ENAS4D, we search for the optimal multi-stage CNN architecture on the ImageNet classification task. ImageNet~\cite{Deng_ImageNet_CVPR2009} contains 1,281,167 samples of 1,000 classes for training and 50,000 samples for validation. We use the original
validation set for testing and sample 10,000 images from training set for evaluating the metric and setting the thresholds. We also evaluate different aspects of the searched network, which are presented in the discussion part. 

\subsection{Experiment Settings}

\subsubsection{Model setup} 
In our experiments, we use the same once-for-all network in~\cite{Cai_onceforall_ICLR20}. We initialize the network with the pre-trained weight. The metric predictor is a 3-layer perceptron that has 400 hidden units in each layer.

\subsubsection{Sample settings}
The number of stages $S$ is set to 3, the input resolution pool of the network is set to [128, 144, 160, 176, 192, 208, 224], the depth pool of the block group is set to [2 ,3 ,4], the kernel size pool of the depth-wise convolution is set to [3, 5, 7]. In each residual block, the cumulative ratio of the number of intermediate channels can be select form [1/2, 2/3, 1].

% the ratio of the intermediate channels to once-for-all in the stage 1 is select from [1/2, 2/3] and that of the stages 2 and 3 are select from [1/6, 1/3], and the sum of ratios for the three stages should be restricted to less than or equal to 1. 

% 我们设置多阶段网络的阶段数量为3，网络的输入分辨率池设置为[128,144,160,176,192,208,224]，block group的深度池设置为[2,3,4]，block的depth-wise卷积层的kernel size池设置为[3,5,7]，第一个阶段的中间channels占once-for-all所有channels的比例可选范围为[1/2, 2/3, 1]，第二、三个阶段可选范围为[0,1/6,1/4,1/2]，另外需要限制三个阶段的比例加起来小于等于1。

\subsubsection{Training details}
We use the standard SGD optimizer with Nesterov momentum 0.9 and weight decay 3e-5. The initial learning ratio is set to 4.5e-3 and we use cosine schedule for learning rate decay. The fine-tune from pre-trained network takes 24 epochs with batch size 192 on 3 GPUs. The training takes about 90 GPU hours.

We set $\omega$ to 0.09 for evaluation metric. And we use the Adam optimizer for the training of metric predictor for 30 epochs. We use the RMSE between the predicted metric and estimated metric as the loss. The learning rate is set to 1e-4 and the weight decay is set to 1e-5.

\subsection{Search Results on ImageNet}

\subsubsection{Searched networks under different cost target}

% \usepackage{booktabs}
\begin{table}[htb]
\centering
\caption{Evaluations on the searched multi-stage CNN architectures for dynamic inference. The average top-1 accuracy on ImageNet, average computational cost in MACs and the cumulative computational cost for each stage are reported.}

\begin{tabular}{@{}cccccc@{}}
\toprule
\multirow{2}{*}{Network} & \multirow{2}{*}{\begin{tabular}[c]{@{}c@{}}Average\\ top-1\end{tabular}} & \multirow{2}{*}{\begin{tabular}[c]{@{}c@{}}Average\\ MACs\end{tabular}} & \multicolumn{3}{c}{Cumulative MACs} \\ \cmidrule(l){4-6} 
                         &                                                                          &                                                                         & Stage 1    & Stage 2    & Stage 3   \\ \midrule
A                        & 70.5                                                                     & 92                                                                      & 75         & 148        & 240       \\
B                        & 70.8                                                                     & 123                                                                     & 90         & 175        & 218       \\
C                        & 72.1                                                                     & 153                                                                     & 103        & 149        & 237       \\
D                        & 74.4                                                                     & 185                                                                     & 141        & 201        & 359       \\ \bottomrule
\end{tabular}
\label{table:compare}
\end{table}

We set the target computation cost to 90M MACs (ENAS4D-A), 120M MACs(ENAS4D-B), 150M MACs (ENAS4D-C) and 180M MACs (ENAS4D-D) in evaluation metric, and perform the architecture search to get the multi-stage CNNs.
Table~\ref{table:compare} reports the performance of the searched multi-stage networks using ENAS4D. 
With different ${\rm COST_{target}}$, ENAS4D is able to sample a suitable multi-stage CNN architecture from once-for-all network. 
The cumulative MAC of the network increases as the stage increases. The computation costs of the stage are much smaller than that of the third stage and that of the second stage is near the average of the first and third stages. 
This shows that ENAS4D can select the proper sub-network for different stages, so that each stage can extract a reasonable number of features to process samples of different difficulty. The detailed analysis for different stages is located at the discussion part.

% 在不同的的target情况下，ENAS4D都能从once-for-all network里面sample到合适大小的多阶段网络
% 网络的累计MAC随着stage的增加而增加
% 第一个阶段的计算量显著少于三个阶段的计算量。
% 第二个阶段的计算量位于第一个和第三个阶段的均值附近
% 这显示了我们的ENAS4D能很好地对选择三个阶段的计算量，使得每个阶段都能提取合理数量的特征来处理不同难度的样本。

\subsubsection{Comparing with other methods for dynamic inference}

\begin{figure}[tbp] % h:here 当前位置 % b bottom % t top % p 浮动
    \centering
    \includegraphics [width=1\linewidth]{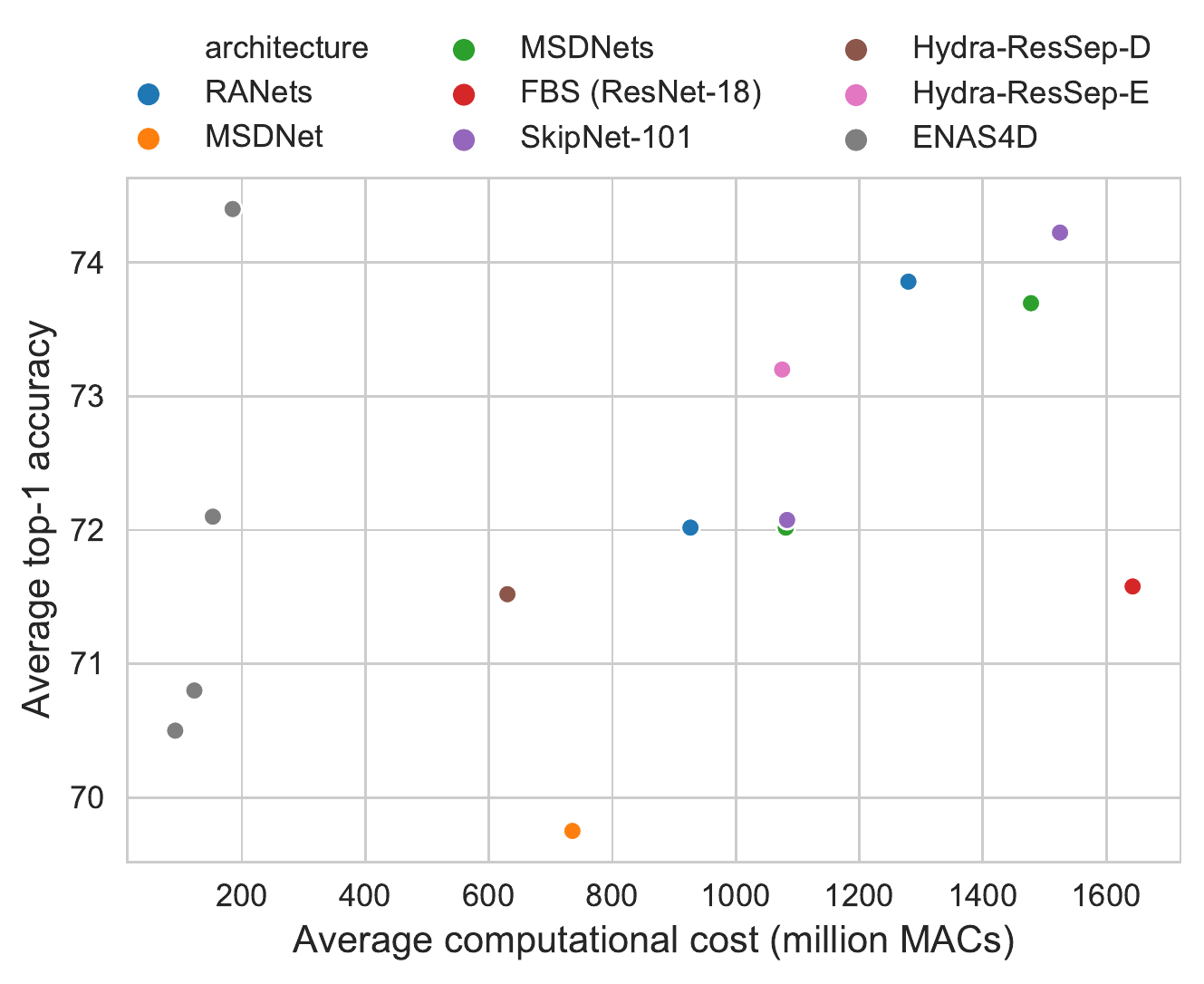}
    \caption{Comparing ENAS4D with other methods for dynamic inference. We plot the points of (the average computation cost, the average top-1 accuracy) on test set of ImageNet. }
    \label{fig:compare_dynamic}
\end{figure}

Figure~\ref{fig:compare_dynamic} comparing the performance of the multi-stage CNNs generated with ENAS4D with other methods for dynamic inference, including RANet~\cite{YangRANetCVPR20}, MSDNet~\cite{Huang2018MSDNet}, HydraNet~\cite{Mullapudi_HydraNet_CVPR2018} SkipNet~\cite{Wang2018SkipNet} and FBS~\cite{DBLP:conf/iclr/FBS19}.
The networks search by ENAS4D are substantially more accurate than other networks for dynamic inference with the same amount of computation cost. 
% The experiments show that the ENAS4D-A is  state-of-the-art RANets

\subsubsection{Time cost}

% \usepackage{booktabs}
\begin{table}[htb]
\caption{Time cost for each step. The data of pre-train once-for-all network is from \cite{Cai_onceforall_ICLR20}.}
\begin{tabular}{@{}ccc@{}}
\toprule
Step                                                                                     & Reusable & Cost           \\ \midrule
Pre-train once-for-all network                                                           & $\surd$        & 2000 GPU hours \\
Train for dynamic inference                                                              & $\surd$        & 90 GPU hours   \\
\begin{tabular}[c]{@{}c@{}}Build the confidence \\ and correctness database\end{tabular} & $\surd$        & 100 GPU hours  \\
Train metric predictor                                                                   & ×        & $<$ 10 min         \\
Evolutionary search                                                                       & ×        & $<$ 3 min          \\ \bottomrule
\end{tabular}
\label{table:time_cost}
\end{table}

Table~\ref{table:time_cost} demonstrates the time cost of each step of ENAS4D.
Training once-for-all for dynamic inference from scratch takes about 2090 GPU hours, which only needs to be executed once. 
The testing of the confidence and correctness for the 18K sampled multi-stage CNNs takes 100 GPU hours. Since we store the results in the database, this time-consuming testing also needs to be executed once.
Comparing to the previous steps, the time for training of metric predictor and evolutionary searching is negligible. We can efficiently search for different multi-stage CNN architectures under different constraint of computation cost.

% 从零开始训练once-for-all for dynamic inference总共需要花费 (2000+90) GPU hours的时间，这个训练只需要进行一次。

% 表格展示了每一个步骤的时间开销。
% 由于我们将结果存储在数据库中，这个耗时的测试仅需执行一次

\subsection{Discussion}

% \subsubsection{Prediction accuracy of metric predictor}

\subsubsection{Difficulty distribution of test dataset}

% \usepackage{multirow}
\begin{table}[htb]
\centering
\caption{Top-1 accuracy and fractions of samples in test set of ImageNet that exit from each stage.}

\begin{tabular}{@{}cccccc@{}}
\toprule
\multicolumn{2}{c}{Network}            & A      & B      & C      & D      \\ \midrule
\multirow{2}{*}{Stage 1} & top-1     & 79.4\% & 85.7\% & 94.6\% & 89.5\% \\
                         & Fractions & 80.3\% & 67.2\% & 44.3\% & 64.7\% \\ \midrule
\multirow{2}{*}{Stage 2} & top-1     & 37.1\% & 49.3\% & 70.5\% & 57.1\% \\
                         & Fractions & 17.3\% & 21.6\% & 27.9\% & 20.9\% \\ \midrule
\multirow{2}{*}{Stage 3} & top-1     & 15.6\% & 23.1\% & 38.2\% & 31.8\% \\
                         & Fractions & 2.3\%  & 11.2\% & 27.9\% & 14.4\% \\ \bottomrule
\end{tabular}
\label{table:difficulty}
\end{table}

In Table~\ref{table:difficulty}, we give the statistics of all the samples in the test set of ImageNet.
We observed that more than 60\% of the samples are exit form the first two stages, which means most of the images are simple samples. The accuracy of the first stage is much higher than the later stages, which indicates that the small networks in early stage can easily classify those samples.

% \subsubsection{Budgeted batch classification}

% \begin{figure}[tbp] % h:here 当前位置 % b bottom % t top % p 浮动
%     \centering
%     \includegraphics [width=0.9\linewidth]{images/budget.pdf}
%     \caption{Top-1 accuracy of budgeted batch classification on ImageNet.}
%     \label{fig:budget}
% \end{figure}

% The budgeted batch classification~\cite{Huang2018MSDNet} is a type of inference method for multi-stage networks, where a fixed amount of computation is available to process a batch of $M$ samples. 
% The result on the test set of ImageNet is shown in Figure~\ref{fig:budget}. The ENAS4D outperform RANets and MSDNets across all budgets. For low computational budget, the ENAS4D- ....
% 在我们的实验中，通过给每一个阶段设置合适的阈值来进行动态推理以满足budgeted batch classification的要求。具体来说，我们将COST_avg的值设置为限制阈值，然后使用网络搜索找到最优的omega设定和阈值设定。

% 

\subsubsection{Visualization}

\begin{figure}[tbp] % h:here 当前位置 % b bottom % t top % p 浮动
    \centering
    \includegraphics [width=0.85\linewidth]{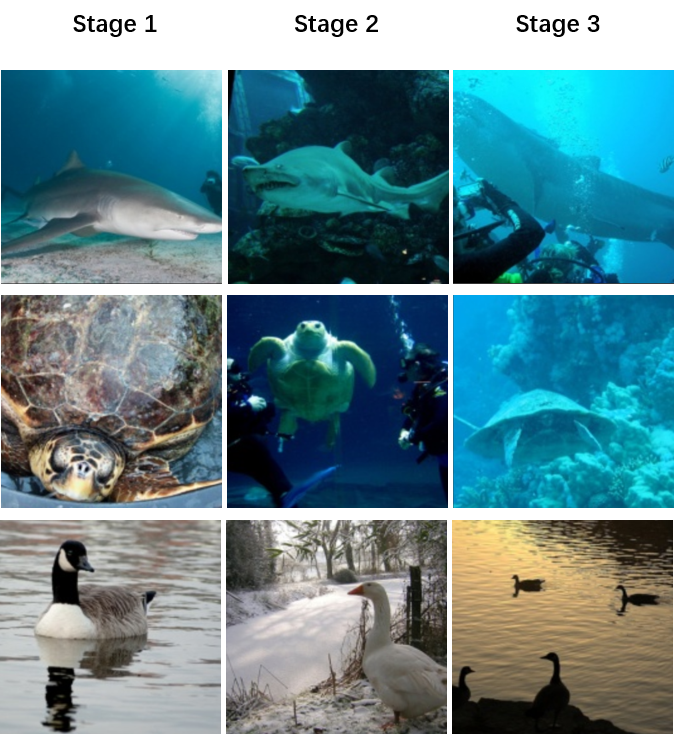}
    \caption{Visualization of ImageNet samples exit from different stages.}
    \label{fig:visualize}
\end{figure}

Figure~\ref{fig:visualize} illustrates the samples exit from different stages the test set of ImageNet. We can see from the figure that the sample exit from different stages have different recognition difficulties. With the rise of the stage, the difficulty of recognition is also rising, which also validates the intuition that easy samples can be classified using fewer computations.

% 我们可以从图中看到，从不同阶段退出的图片有不同的识别难度。随着阶段的上升，识别难度也在不断的上升。
% \subsubsection{Searched architecture}

\subsubsection{Deploy on real-world devices}

\begin{table}[htb]
\centering
\caption{Evaluations on the searched multi-stage CNN architectures for dynamic inference. The average top-1 accuracy on ImageNet and average latency on the desktop and the GPU server.}
\begin{tabular}{@{}cccc@{}}
\toprule
\multirow{2}{*}{Platform}                                             & \multirow{2}{*}{Target latency} & \multirow{2}{*}{Average top-1} & \multirow{2}{*}{Average latency} \\
                                                                      &                                 &                                &                                  \\ \midrule
\multirow{3}{*}{Desktop}                                              & 10ms                            & 74.2                           & 9.8ms                            \\
                                                                      & 8ms                             & 72.9                           & 8.0ms                            \\
                                                                      & 6ms                             & 72.2                           & 6.4ms                            \\ \midrule
\multirow{3}{*}{\begin{tabular}[c]{@{}c@{}}GPU\\ Server\end{tabular}} & 30ms                            & 74.2                           & 30.2ms                           \\
                                                                      & 25ms                            & 73.3                           & 25.4ms                           \\
                                                                      & 20ms                            & 72.8                           & 20.6ms                           \\ \bottomrule
\end{tabular}
\label{table:intel_cpu}
\end{table}

In real-world deployment, multi-stage CNNs needs to run on specific hardware. 
However, the computational efficiency of each type of hardware for different network architecture is different. So we should search for the optimal multi-stage CNN architecture for each specific hardware. 
Therefore, it is necessary to change the ${\rm COST_{avg}}$ and ${\rm COST_{target}}$ from MACs to latency on specific hardware. Since we have established a database of prediction confidence and correctness before, we can quickly generate data and train a new metric predictor. We chose a desktop (Intel i7-6700K) and a GPU server (Nvidia GTX 1080ti) for experimentation. We first build a database for the hardware latency of operations under different input conditions. 
We only need to look up the table to get the latency of different stages of a multi-stage architecture. Table~\ref{table:intel_cpu} demonstrates the result on the desktop and the GPU server.

% 在实际的部署中，神经网络需要运行在特定的硬件上。然而，每一种硬件对于不同网络配置的计算效率都是不同的，我们需要为每一种不同的硬件搜索最优的多阶段网络结构。因此需要将计算metric的指标从MACs换成实际运行latency。
% 由于于我们之前已经建立了prediction confidence and correctness数据库，我们能快速地生成数据并训练新的metric predictor。

The inference of the generated multi-stage CNNs does not introduce irregular computations or complex controllers. Thus the generated model can be easily deployed on various hardware devices using existing deep learning frameworks. 
Moreover, our method is orthogonal to quantization methods, which can further reduce the computational cost of the multi-stage CNNs. All these properties of our method imply a wide range of application scenarios where the efficient CNN inference is desired.

\section{Conclusion}
In this paper, we present a general framework called ENAS4D, for efficiently search for the multi-stage CNN architectures for dynamic inference. 
In contrast to previous methods, our method comes with three advantages: (1) Our method can simultaneously search for different architecture for different stages. 
(2) Our method generate a large search space including numbers of layers, kernel sizes, numbers of channels and resolution of input image.
(3) Our method has high search efficiency by using the technique of once-for-all.
The experiments on ImageNet classification benchmark demonstrate the effectiveness of the proposed method. The generated multi-stage CNN architecture can consistently outperform the previous methods for dynamic inference.

{\small
\bibliography{references}
}

\end{document}